\documentclass[]{style/ceurart}
\usepackage{gensymb}
\usepackage{graphicx}
\usepackage{booktabs}
\usepackage{multirow}
\usepackage{subcaption}
\usepackage{listings}
\begin{document}
\copyrightyear{2026}
\copyrightclause{Copyright for this paper by its authors.
  Use permitted under Creative Commons License Attribution 4.0
  International (CC BY 4.0).}

\conference{CLEF 2026: Conference and Labs of the Evaluation Forum, September 21-24, 2026, Jena, Germany}

\title{Multi-Scale ViT Inference with Habitat-Fit Priors and kNN Retrieval for Multi-Species Plant Identification}

\author[1]{Alper Erten}[
orcid=0009-0007-3417-0584,
email=aerten3@gatech.edu,
url=https://alpererten.com
]
\cormark[1]

\author[1]{Murilo Gustineli}[
orcid=0009-0003-9818-496X,
email=murilogustineli@gatech.edu,
url=https://murilogustineli.com,
]
\cormark[1]

\author[1]{Adrian Cheung}[
orcid=0009-0006-8650-4550,
email=acheung@gatech.edu,
]

\address[1]{Georgia Institute of Technology, North Ave NW, Atlanta, GA 30332}
\cortext[1]{Corresponding author.}

\begin{abstract}
    This paper describes DS@GT ARC's third-place solution to the PlantCLEF 2026 challenge on multi-species plant identification in vegetation quadrat images, where systems must predict every species present in high-resolution ($\approx\!3000\!\times\!3000$ pixel) plot photographs while training only on single-label images of individual plants.
    The pipeline is built around a fine-tuned DINOv2 ViT-L/14 classifier applied over a multi-scale tile decomposition of each quadrat, with per-tile predictions blended with a FAISS kNN retriever and post-processed by source-aware temporal fusion across repeated plot visits, a habitat-fit demotion that injects geographic and altitude priors from the training data, and a South-Western Europe geographic mask.
    Habitat-fit demotion and multi-scale aggregation are the largest individual contributors in the ablations.
    Two complementary training-centric directions, a cross-region transformer with noisy-student distillation on the LUCAS dataset and a label-as-query transformer decoder over synthetic CLS-domain pseudo-quadrats, yielded null results.
    An inference-time augmentation with instance-aware segmentation crops also did not improve performance.
    The selected submission reaches a private-leaderboard macro-$F_1$ of $0.43902$ (third place; public $0.51096$); an unselected configuration of the same pipeline scored above $0.45$ on the private set. Code: \url{https://github.com/dsgt-arc/plantclef-2026}.
\end{abstract}


\maketitle

\section{Introduction}
\label{sec:intro}

The PlantCLEF task \cite{plantclef2026} within the LifeCLEF lab \cite{lifeclef2026} 
at the Conference and Labs of the Evaluation Forum (CLEF) asks competitors to identify every plant species present in high-resolution ($\approx\!3000\!\times\!3000$~pixel) top-down photographs of $0.5\!\times\!0.5$~m vegetation quadrats placed on the ground by botanists.
Such quadrat inventories are central to standardized biodiversity assessment, long-term ecological monitoring, and large-scale field surveys;  automating them would allow specialists to extend the temporal and spatial coverage of ecological studies  and enable non-expert citizen scientists to contribute to monitoring programs \cite{plantclef2025overview}.
The 2026 edition reuses the 2025 dataset and Kaggle platform as a second-round benchmark  to consolidate methodological progress on this challenge \cite{plantclef2026}.

There are two main challenges in the task.
First, there is a severe single$\to$multi-label \emph{domain shift}: the only labeled training data consists of approximately $1.4$~million Pl@ntNet images of individual plants centered on a single specimen (often a close-up of a single organ such as a flower, leaf, or fruit), while every test image is a cluttered ground-cover scene containing multiple co-occurring species at varied scales and phenological stages \cite{plantclef2024overview, plantclef2025overview}.
Second, the test images are far too large to feed directly into a standard $518\!\times\!518$ Vision Transformer input, forcing competitors to adopt some form of tiling and per-tile aggregation.
The evaluation metric is a macro-averaged per-sample $F_1$ score, further averaged across transects to avoid bias from over-sampled sites \cite{plantclef2025overview}.

The pipeline is built around a fine-tuned DINOv2 ViT-L/14 backbone applied over a multi-scale tiling decomposition ($3{\times}3$, $4{\times}4$, $5{\times}5$, $6{\times}6$; 86 tiles per image). Per-tile predictions are aggregated via max pooling at temperature $T=1.5$, preserving strong local evidence across scales.
To complement the classifier, a FAISS-based kNN retrieval module over $\sim\!860{,}000$ ArcFace-trained embeddings is incorporated, blended at the tile level.
Beyond model architecture, dataset structure is also leveraged by introducing per-source temporal fusion across repeated visits, followed by a habitat-fit demotion step that injects geographic and altitude priors derived from the training data.
A global South-Western Europe geographic mask and a two-stage admission rule further refine the final predictions.

The team also explored three complementary directions that did not improve results: a cross-region transformer trained with noisy-student distillation on the unlabeled 212K-image LUCAS dataset \cite{dandrimont2022lucas} (\S\ref{sec:methodology}), a label-as-query transformer decoder (TileQ-Decoder) over per-tile DINOv2 CLS embeddings trained on synthetic CLS-domain pseudo-quadrats (\S\ref{sec:tileq-decoder}), and an inference-time augmentation using instance-aware crops from a SAM 3 segmentation model \cite{sam3}.
All three yielded null results, consistent with the strength of the aggregation-based baseline.

On the official private leaderboard (computed on the $\approx\!89\%$ holdout portion of the test set), the team's final selected submission reached a macro-averaged $F_1$ of $\mathbf{0.43902}$, placing third overall. 
The corresponding public-leaderboard score, on the remaining $\approx\!11\%$ of test data, was $\mathbf{0.51096}$.
During the competition, intermediate configurations of the same ensemble pipeline achieved private-leaderboard scores above $0.45$; those configurations were not selected as one of the five final submissions before the deadline.
Both the selected and unselected scores represent a substantial improvement over the team's 2025 result of $0.348$ private-leaderboard $F_1$ \cite{gustineli2025tile}  and over the best 2024 result of $0.287$ \cite{plantclef2024overview}.
All code, configuration files, and reproducibility scripts are publicly available at \url{https://github.com/dsgt-arc/plantclef-2026}.

\section{Related Work}
\label{sec:related-work}

\paragraph{The PlantCLEF series.}
PlantCLEF has run annually since 2011 as part of LifeCLEF, beginning with a 71-species classification task on the ImageCLEF 2011 plant identification challenge \cite{joly2019biodiversity, plantclef2023}.
From 2017 through 2023 the task was a single-label, global-scale species identification problem evaluated on isolated-plant images, with editions covering progressively more species: $10{,}000$ species in 2017 (illustrated by a total of $1.1$M images) \cite{plantclef2017}  and $80{,}000$ species  in both the 2022 and 2023 editions \cite{plantclef2022, plantclef2023}.
The dominant approaches in those editions were CNN ensembles and, increasingly, Vision Transformers fine-tuned on noisy Pl@ntNet data.

The 2024 edition \cite{plantclef2024overview} fundamentally changed the task into a weakly-supervised multi-label problem on high-resolution vegetation plots, addressing the single$\to$multi-label domain-shift problem that defines the present work.
The 2025 edition \cite{plantclef2025overview} kept the same problem formulation but introduced a new, more diverse $2{,}105$-image quadrat test set  and added a complementary $212{,}782$-image unlabeled pseudo-quadrat training set derived from the LUCAS Cover Photos archive \cite{dandrimont2022lucas} to facilitate self-supervised domain adaptation.
The 2026 edition reuses the 2025 datasets as a second round. 

\paragraph{Top approaches in PlantCLEF~2024 and 2025.}
All three teams that produced 2024 working notes adopted a \emph{tiling inference} pipeline on top of the provided ViT-B/14 DINOv2 checkpoint fine-tuned on the 2024 training data \cite{plantclef2024overview}.
The winning Atlantic team \cite{foy2024atlantic} combined three tile configurations, used Segment Anything \cite{kirillov2023sam}  for non-plant rejection, and aggregated predictions across repeated visits to the same plot, reaching a sample-$F_1$ of $0.2873$.
NEUON~AI \cite{chulif2024patch} proposed RICAP-inspired \cite{takahashi2020ricap}  composite training images and Bayesian Model Averaging across tile predictions.
The DS@GT-LifeCLEF group \cite{gustineli2024multilabel} explored linear classifiers on top of DINOv2 embeddings with Spark-based distributed preprocessing.

In 2025, the top private-leaderboard submissions by Espitalier \cite{espitalier2025preprocessing}, DS@GT \cite{gustineli2025tile}, and Chlorophyll Crew \cite{herasimchyk2025multi} all kept the same DINOv2 backbone and refined the tiling/aggregation pipeline with (i) careful JPEG re-encoding aligned to the training-set chroma subsampling,  (ii) multi-scale tile grids, (iii) taxonomy-aware multi-head classification, and (iv) ecological priors derived from GBIF and Ellenberg indicator values \cite{dengler2023ecological}.
The DS@GT 2025 entry, which the present work extends, used a $4\!\times\!4$ tile grid  with PaCMAP + K-Means visual-cluster priors and a geolocation mask, finishing second on the private leaderboard with $F_1 = 0.348$ \cite{gustineli2025tile}.

\paragraph{DINOv2 for plant identification.} DINOv2 \cite{oquab2023dinov2} is a self-supervised Vision Transformer trained on the curated $142$-million-image LVD-142M dataset. 
Oquab et al.\ train a 1B-parameter ViT model and distill it into a series of smaller models that reportedly outperform general-purpose features including OpenCLIP on a broad range of image-and pixel-level benchmarks \cite{oquab2023dinov2}.
The addition of register tokens \cite{darcet2024vision} stabilizes attention maps and slightly improves classification.
Every PlantCLEF 2024 and 2025 working note used DINOv2 features in some form, and the present work follows that same convention.

\section{Dataset and Model}
\label{sec:dataset-model}

\subsection{Dataset}

\begin{figure}[!htbp]
    \centering
    \includegraphics[width=0.8\textwidth]{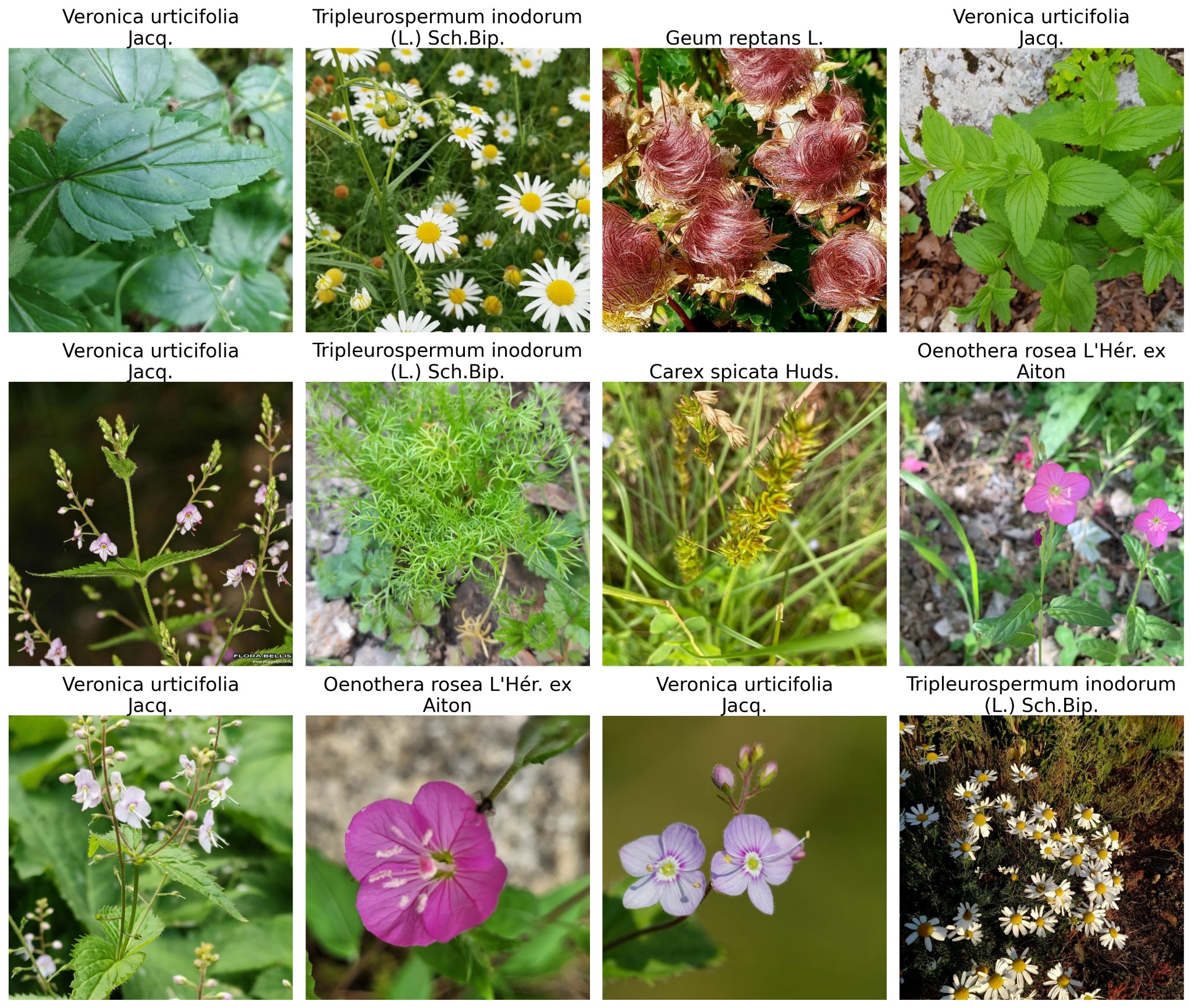}
    \caption{
    Twelve single-label training images displaying the following six species: \textit{Veronica urticifolia Jacq., Tripleurospermum inodorum (L.) Sch.Bip., Geum reptans L., Carex spicata Huds., Oenothera rosea L'Hér. ex Aiton, Lamium bifidum Cirillo}.
    }
    \label{fig:train-single-label}
\end{figure}

PlantCLEF~2026 reuses the PlantCLEF~2024/2025 datasets \cite{plantclef2024overview, plantclef2025overview}. 
The training set is a subset of the Pl@ntNet  collaborative database covering the flora of South-Western Europe: $1{,}408{,}033$ images  of $7{,}806$ vascular plant species  across $1{,}446$ genera and $181$ families, supplemented with trusted-label GBIF images for under-represented species.
Each training image is a single-label photograph of an isolated specimen (often a close-up of a flower, leaf, or fruit), with a minimum or maximum of $800$ pixels  on the corresponding side depending on the released variant (Figure \ref{fig:train-single-label}).
The data is provided pre-organized into species-specific subfolders and a predefined train/val/test split for individual-plant identification ($1{,}308{,}899$ / $51{,}194$ / $47{,}940$ images  respectively).
A complementary $212{,}782$-image unlabeled pseudo-quadrat dataset derived from the LUCAS Cover Photos 2006–2018 archive \cite{dandrimont2022lucas} is available for self-supervised domain adaptation; this dataset was not used in the final pipeline.

\begin{figure}[!htbp]
    \centering
    \includegraphics[width=0.8\textwidth]{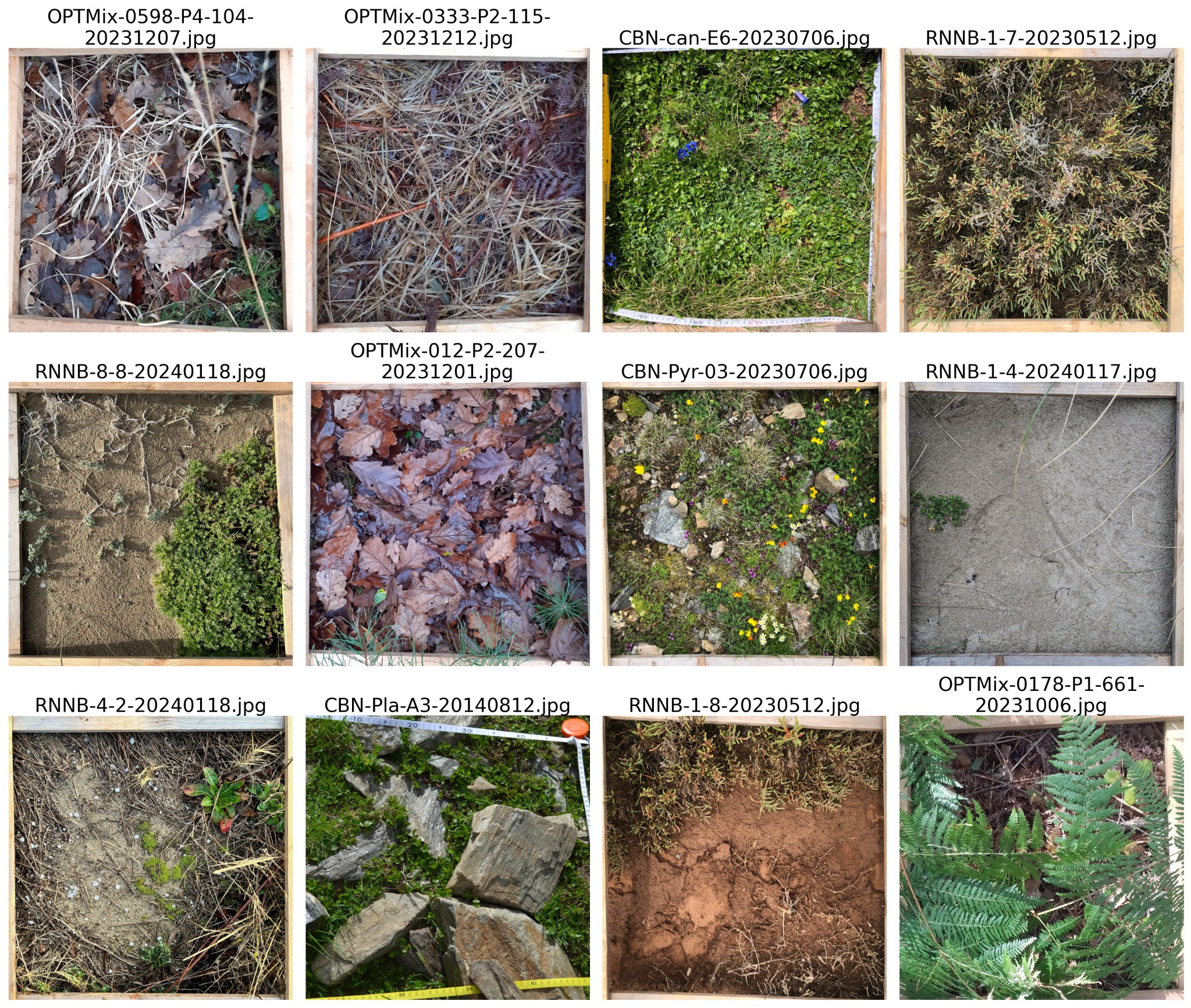}
    \caption{
    Subset of twelve test set images showcasing the significant domain shift between different quadrats.
    }
    \label{fig:test-multi-label}
\end{figure}

The challenge test set is identical to the 2025 set: $2{,}105$ high-resolution  (typically $\approx\!3000\!\times\!3000$~pixel) top-down photographs of $0.5\!\times\!0.5$~m quadrats produced by botanical experts across Pyrenean, Mediterranean,  and South-Western European temperate floras (Figure \ref{fig:test-multi-label}).
Test quadrats are organized into transects (spatially-structured sampling sites), and many transects contain multiple visits to the same plot over months or years.
Roughly $75.8\%$ of the test quadrats belong to a multi-visit transect, which the pipeline exploits through temporal prior fusion.

\paragraph{Evaluation metric.} The official metric is a hierarchical macro-averaged sample $F_1$ score with set-based per-image comparison \cite{plantclef2025overview}.
For each test image $i$, precision and recall are computed by treating the predicted and true species sets as binary multi-labels, and an image-level $F1_i$ is computed in the usual way.
Image-level scores are first averaged within each transect to remove bias from differently-sized transects, and these per-transect averages are then averaged across all transects:
\begin{equation}
    F1_{\text{avg-transect}} = \frac{1}{T}\sum_{k=1}^{T} \left(\frac{1}{Q_k}\sum_{j=1}^{Q_k} F1_j^{(k)}\right)
\end{equation}
where $T$ is the number of transects and $Q_k$ the number of quadrats in transect~$k$.
The Kaggle leaderboard reports this final transect-averaged sample $F_1$ score, split into a public set on $\approx\!11\%$ of the test data and a private set on the remaining $\approx\!89\%$ that determines the official ranking.

\subsection{Pre-trained model}
The competition organizers provide two PlantCLEF-2024 fine-tuned Vision Transformer checkpoints through Zenodo \cite{goeau2024pretrainedmodels} and HuggingFace.
Both are based on the ViT-B/14 architecture with four register tokens \cite{darcet2024vision} originally pre-trained by self-supervised DINOv2  on the LVD-142M dataset of  142 million images \cite{oquab2023dinov2} (the underlying \texttt{timm} model identifier is \texttt{vit\_base\_patch14\_reg4\_dinov2.lvd142m}).
The model has $86$~million parameters  in the distilled ViT-B/14 backbone \cite{oquab2023dinov2},  processes $518 \times 518$ inputs  as $37 \times 37$ patches of size $14$, and produces $768$-dimensional CLS-token embeddings.

The first checkpoint, \texttt{vit\_base\_patch14\_reg4\_dinov2\_lvd142m\_pc24\_onlyclassifier} (ViTD2PC24OC), trains only  a $7{,}806$-way linear classification head on top of the frozen  LVD-142M backbone for $92$ epochs at batch size $1{,}280$ per GPU and a learning rate of $0.01$ on an $8 \times $A100 node, reaching a top-1~/~top-5 single-plant accuracy of $63.69\%$~/~$83.88\%$.
The second checkpoint, \texttt{vit\_base\_patch14\_reg4\_dinov2\_lvd142m\_pc24\_onlyclassifier\_then\_all} (ViTD2PC24All), initializes from ViTD2PC24OC and then fully fine-tunes both the backbone and head  for another $92$ epochs at a much lower learning rate  ($8\!\times\!10^{-5}$) and per-GPU batch size $144$,  reaching a single-plant top-1~/~top-5 of $75.91\%$~/~$92.82\%$ \cite{plantclef2024overview, goeau2024pretrainedmodels}.
Both models are trained with cross-entropy loss; the produced ViTD2PC24All backbone is no longer SSL-aligned.

Following the 2024 and 2025 working-notes consensus, the pipelines build on this second (fully fine-tuned) checkpoint, referred to as ViTD2PC24All throughout the paper.
ViTD2PC24All is used directly as the backbone for the TileQ-Decoder experiments (\S 4.7), and as the ViT-B reference checkpoint in the backbone ablation of \S 5.3.
The classifier as well as the frozen embedding extractor for the FAISS retrieval index (an \texttt{IndexFlatIP} structure built over $\approx 860{,}000$ L2-normalized CLS embeddings of the geo-filtered training set; \S 4.3) of the final pipeline, however, is a larger ViT-L/14 DINOv2 backbone fine-tuned on the same PlantCLEF training data using the two-phase classifier recipe of \S 4.1 (Phases 1--2; the retrieval head is Phase~3, Appendix~\ref{app:training}).

\section{Methodology}
\label{sec:methodology}

\begin{figure}[!htbp]
    \centering
    \includegraphics[width=0.9\textwidth]{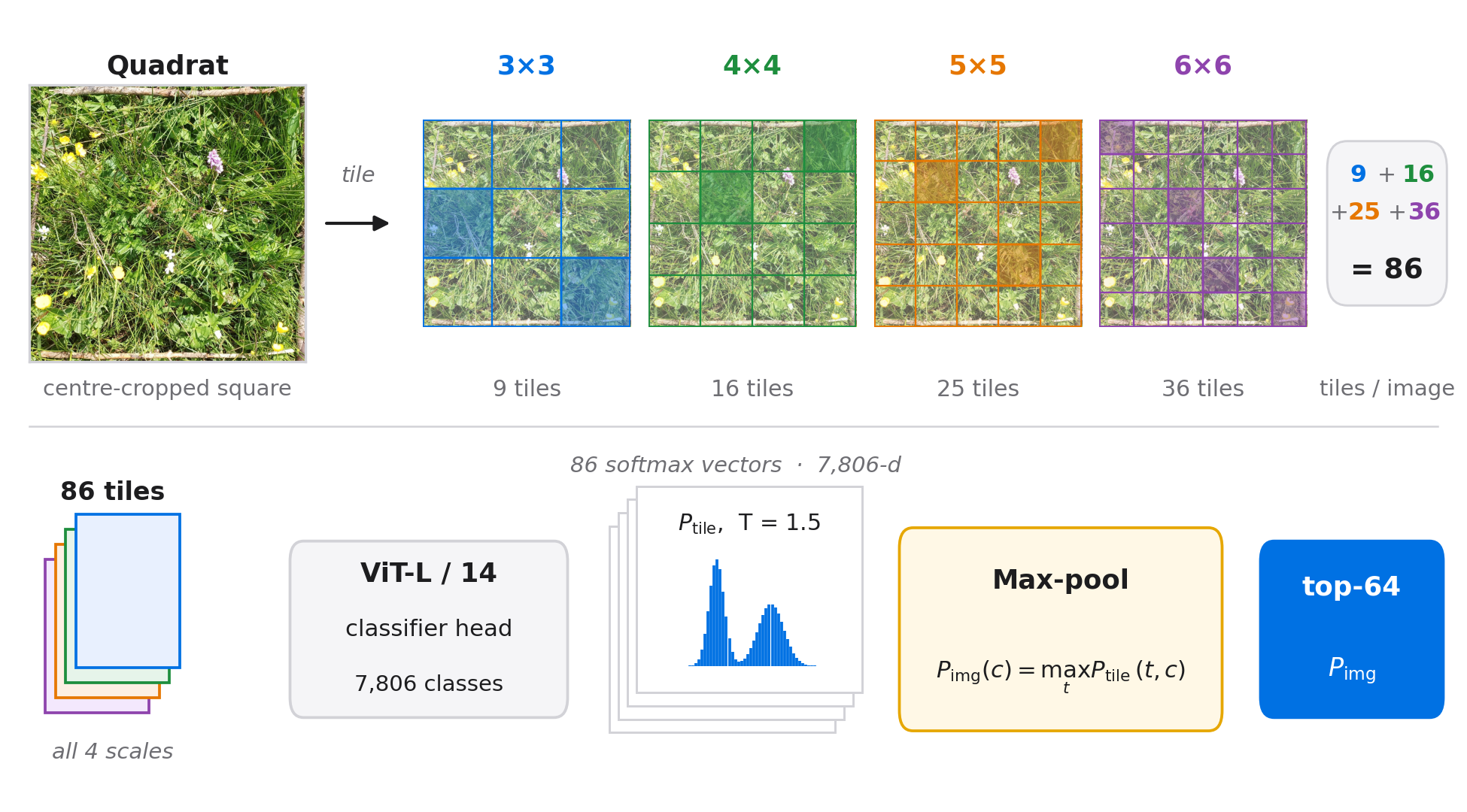}
    \caption{
    Multiscale tile aggregation pipeline. A quadrat image is centre-cropped, tiled at multiple scales, processed by a ViT-L/14 classifier, and aggregated via max-pooling to produce the final per-image prediction.
    }
    \label{fig:multiscale-tiling}
\end{figure}

The final pipeline applies the team-fine-tuned ViT-L (\S\ref{sec:dataset-model}) to a multi-scale tiling of each high-resolution quadrat, blends per-tile predictions with a FAISS-based kNN retriever, and post-processes the aggregated probabilities with source-aware temporal fusion, habitat-fit demotion, and a geographic admission rule.
Figure \ref{fig:multiscale-tiling} gives an overview. The following subsections describe each stage; the rationale and ablations are deferred to Sections \ref{sec:results} and \ref{sec:discussion}.

\paragraph{Development sequence.}
The pipeline was built incrementally. The starting point was a multi-scale tile classifier using the organizer-provided ViT-B/14 checkpoint, which was then replaced with a team-fine-tuned ViT-L/14 backbone for a consistent improvement across all scale configurations (\S\ref{sec:results}). FAISS kNN retrieval was added next and blended at the tile level, followed by per-source temporal fusion to exploit the multi-visit structure of the test set. Habitat-fit demotion and the geographic mask were the final additions and ultimately the largest contributors, as confirmed by the cumulative ablation in Table~\ref{tab:cumulative_ablation}. Three complementary directions (cross-region noisy-student distillation, the TileQ-Decoder, and SAM instance-crop augmentation) were explored in parallel but yielded null results and were not included in the final pipeline.

\subsection{ViT Backbone Scaling and Fine-Tuning}

To further improve the ViT classifier backbone performance, a larger DINOv2 Vision Transformer (ViT-L/14, ~300M parameters) was fine-tuned instead of the ViT-B/14 (~86M parameters) model provided by the PlantCLEF organizers. Following the training strategy described in the PlantCLEF reference pipeline, the classifier backbone was trained in two phases on the 1.4M-image PlantCLEF dataset (a third phase trains the ArcFace/LoRA retrieval head separately; see Appendix~\ref{app:training}).
Phase 1 consists of head warmup at $518\times518$ resolution, where the backbone is frozen and only the 7{,}806-class linear head is trained. Phase 2 unfreezes the full backbone and continues training for 30 epochs using AdamW with layer-wise learning rate decay (0.85), drop-path (0.3), RandAugment, MixUp, CutMix, and EMA (decay 0.9998). Full hyperparameters for all three phases are reported in Appendix~\ref{app:training}.

To better match the input statistics of the pretrained DINOv2 backbone, training-time augmentation includes a JPEG round-trip (quality 85, 4:2:2 chroma subsampling) applied before normalization. The same preprocessing is used at inference time, following a strategy similar to the top-ranked PlantCLEF 2025 submission \cite{espitalier2025preprocessing}, highlighting the importance of preprocessing alignment for performance gain.

\subsection{Multiscale Tile Aggregation}

A high-resolution quadrat (typically $\sim$3000 px on the long side) cannot be processed directly at $518 \times 518$ resolution without losing fine-grained plant cues. Following prior work \cite{plantclef2025overview}, a multi-scale, non-overlapping tiling strategy is adopted. For each test image and scale $s$, we first extract the largest centered square crop, resize it to $(518s) \times (518s)$ using Lanczos interpolation, and apply a JPEG round-trip (quality 85, 4:2:2 chroma subsampling). For non-square images, the center crop discards a strip of pixels along the longer axis. In practice, the test quadrat images are approximately square ($\approx\!3000\!\times\!3000$~px), so the discarded margin is narrow; however, edge vegetation in strongly non-square images may be excluded from the tile pool. Importantly, the crop operates at the image level: individual tiles are not further cropped and cover their respective regions in full, so border vegetation loss is limited to this single squarification step. On the PlantCLEF 2024 single-plant training images, which are already subject-centred, the center crop trims only background border content and does not exclude relevant plant material. The resized image is then partitioned into $s \times s = s^2$ non-overlapping tiles of size $518 \times 518$. Each tile is passed through the ViT classifier to produce a 7,806-dimensional softmax distribution (temperature $T = 1.5$).

The final pipeline uses the scale set $S = \{3, 4, 5, 6\}$, resulting in $9 + 16 + 25 + 36 = 86$ tiles per image. This configuration was selected based on the multi-scale sweep in Section \ref{sec:results} (Table \ref{tab:multiscale_tiling}), where combining multiple scales improved coverage of small and densely clustered species in high-complexity quadrats.
Per-image aggregation is performed via max pooling over tiles. For each class $c$,
\begin{equation}
P_{\text{img}}(c) = \max_{t \in \text{tiles(img)}} P_{\text{tile}}(t, c).
\end{equation}
Max pooling preserves the strongest evidence for each species independently and is approximately invariant to the number of tiles, enabling stable aggregation across multiple scales. Finally, the per-image probability vector is truncated to the top-64 classes (\texttt{image\_topk} = 64), with all remaining classes set to zero prior to downstream fusion.

\subsection{FAISS kNN Image Similarity Retrieval and Ensembling}

To further improve performance beyond the ViT-L classifier, an image-retrieval component based on k-nearest neighbors (kNN) using FAISS is incorporated.

\begin{figure}[!htbp]
    \centering
    \includegraphics[width=0.95\textwidth]{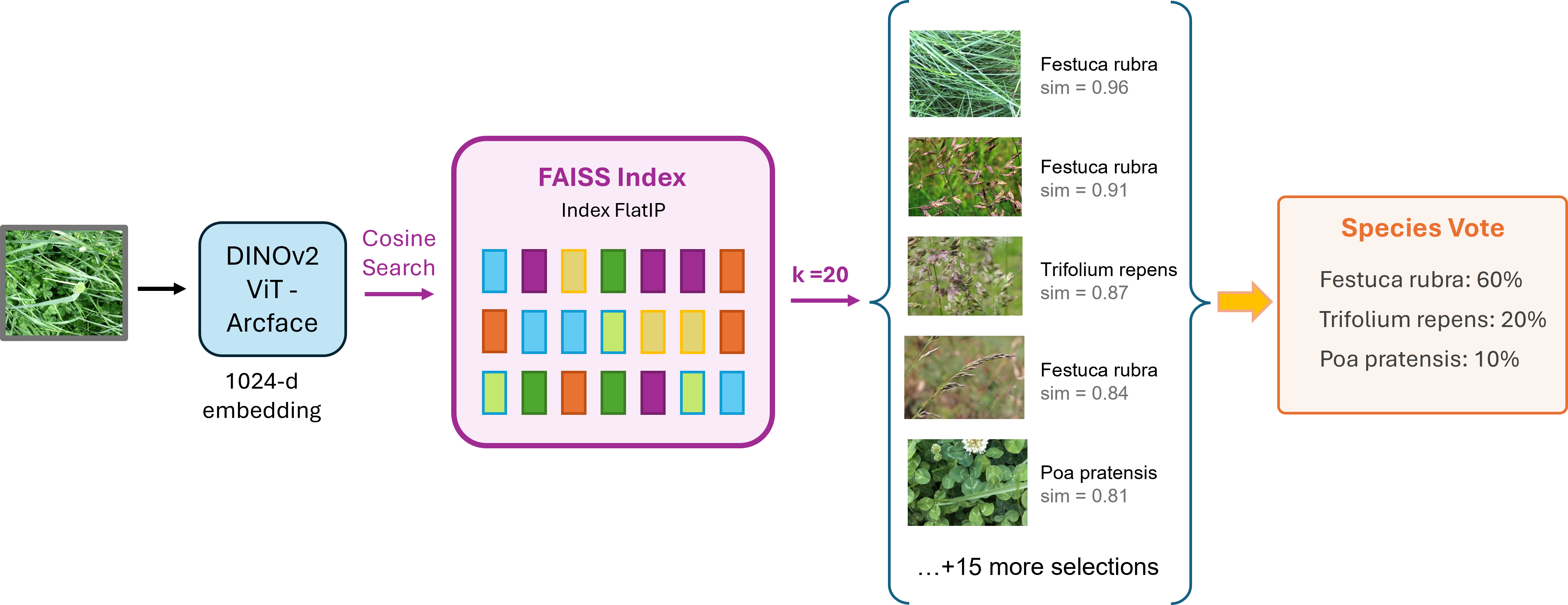}
    \caption{
    FAISS kNN retrieval pipeline: ViT-L + ArcFace embeddings are used to perform cosine similarity search over a FAISS IndexFlatIP index. The top-$k$ nearest neighbors are retrieved, and their similarities are converted into weights to produce a class-level vote distribution, which is later fused with classifier predictions.
    }
    \label{fig:faiss_retrieval}
\end{figure}

Embeddings for the 1.4M PlantCLEF training images are first extracted using the ViT-L backbone. To improve embedding quality for similarity search, we attach a LoRA adapter (rank 16) with an ArcFace margin objective to the backbone. This produces 1,024-dimensional L2-normalized embeddings from the [CLS] token. The adapter is trained for 8 epochs on the same single-label dataset, encouraging embeddings of the same species to cluster in feature space (see Appendix~\ref{app:training} for full training details).

At index construction time, one embedding per training image is extracted, retaining only species observed in South-Western Europe, resulting in approximately 860k vectors. These are indexed using a FAISS \texttt{IndexFlatIP} structure, storing both embeddings and their corresponding class labels.
At inference time, each tile embedding is L2-normalized and queried against the index to retrieve the top-$k=20$ nearest neighbors. Let $\{\text{sim}_i\}$ denote the cosine similarities. Neighbor weights are computed via a temperature-scaled softmax:
\begin{equation}
w_i = \frac{\exp(\text{sim}_i / \tau)}{\sum_j \exp(\text{sim}_j / \tau)}, \quad \tau = 0.07.
\end{equation}
These weights are aggregated by class label to produce a soft 7,806-dimensional retrieval distribution $P_{\text{kNN}}(t, c)$ for each tile $t$. The retrieval distribution is then combined with the classifier output prior to tile aggregation:
\begin{equation}
P_{\text{tile}}(t, c) = \beta \cdot \text{softmax}(z_t(c) / T) + (1 - \beta) \cdot P_{\text{kNN}}(t, c).
\end{equation}

Blending is performed using a visit-aware schedule: $\beta_{\text{multi}} = 0.70$ for multi-visit quadrats and $\beta_{\text{single}} = 1.00$ for single-visit quadrats. Disabling kNN for single-visit images was necessary, as early experiments showed performance degradation of up to 0.08 F1 due to noisy retrievals. Further discussion of this effect is provided in the next section.

\subsection{Temporal Fusion (Per-Source)}

A key property of the 2026 test set is its multi-visit structure, where the same physical plot is photographed on multiple dates. We define the \textit{location identifier} of a quadrat by removing the trailing date suffix from its \texttt{quadrat\_id}; quadrats sharing the same identifier are treated as siblings. Sibling matching is performed strictly at inference time using only the quadrat identifier strings provided in the test set metadata; no ground-truth labels or test-set metadata are used during training or FAISS index construction. The index is built exclusively over the 1.4M PlantCLEF training images, and the visit similarity weights in \texttt{visit\_similarity.json} are derived from ViT-L CLS embeddings of the test tiles, not from any label information. Empirically, 76\% of the 2,105 test images have at least one sibling, making per-location temporal aggregation a central component of the pipeline.

\begin{figure}[!htbp]
    \centering
    \includegraphics[width=0.8\textwidth]{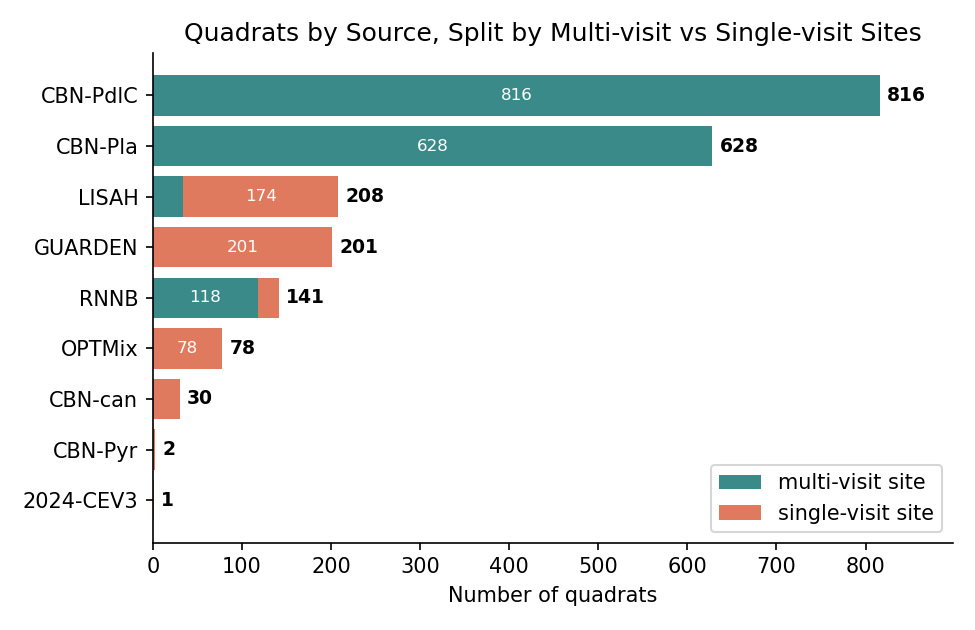}
    \caption{
    Number of quadrats per source, separated by multi-visit and single-visit sites. Most samples belong to multi-visit collections, while several sources (e.g., LISAH, GUARDEN) contain predominantly single-visit data, motivating the use of source-aware temporal fusion.
    }
    \label{fig:visits_by_source}
\end{figure}

The dataset is further stratified by source, where the leading prefix of \texttt{quadrat\_id} identifies one of eight collection sources: \texttt{RNNB}, \texttt{CBN-PdlC}, \texttt{CBN-Pla}, \texttt{CBN-can}, \texttt{GUARDEN}, \texttt{LISAH}, \texttt{OPTMix}, and \texttt{2024-CEV3}. Each source corresponds to a distinct ecological regime (e.g., coastal salt marsh, alpine summit, sub-alpine, Mediterranean, agricultural), leading to substantial variation in species composition, image count, and temporal sampling. Figure \ref{fig:visits_by_source} shows the distribution of quadrats across sources, separated into multi-visit and single-visit collections. As a result, all priors and aggregation steps applied downstream of the classifier are defined on a per-source basis to account for these distributional differences.

For each quadrat $q$ with at least one sibling at the same location, a location prior is constructed by aggregating predictions from its siblings and mix it into the per-quadrat prediction:
\begin{equation}
P_{\text{fused}}(q, c) = \alpha \, P_{\text{img}}(q, c) + (1 - \alpha) \, P_{\text{loc}}(L(q), c),
\end{equation}
where $L(q)$ denotes the location of $q$, and the aggregator $P_{\text{loc}}$ is defined per source.
Two aggregation strategies are considered:
\begin{align}
P_{\text{loc}}^{\text{max}}(L, c) &= \max_{q' \in \text{siblings}(L)} P_{\text{img}}(q', c), \\
P_{\text{loc}}^{\text{sim}}(L, c) &= \sum_{q'} w(q, q') \, P_{\text{img}}(q', c),
\end{align}
with similarity weights
\begin{equation}
w(q, q') = \frac{\exp\left(\cos(e_q, e_{q'}) / \tau_v\right)}{\sum_{q''} \exp\left(\cos(e_q, e_{q''}) / \tau_v\right)},
\quad \tau_v = 0.10.
\end{equation}
Here, $\cos(\cdot, \cdot)$ is computed on the per-quadrat mean of the ViT-L [CLS] tile embeddings. The weights $w(q, q')$ are precomputed and stored in a json file, \texttt{visit\_similarity.json}.

The max aggregator is used for all sources except \texttt{RNNB}. For \texttt{RNNB}, a coastal sand-dune source with high temporal variability, we use the similarity-weighted aggregator. In this setting, max pooling can produce overconfident predictions by equally weighting visually dissimilar visits (e.g., lush summer vs.\ bare winter). In contrast, the similarity-weighted approach emphasizes visually consistent siblings. Section~5 shows that extending this strategy to other sources degrades performance. The mixing coefficient $\alpha$ is also source-dependent: $\alpha_{\text{RNNB}} = 0.30$ and $\alpha_{\text{default}} = 0.15$ for all other sources. Single-visit quadrats (approximately 24\% of the test set) do not apply temporal fusion.

\subsection{Habitat-Fit Demotion}

\begin{figure}[!htbp]
    \centering
    \includegraphics[width=0.85\textwidth]{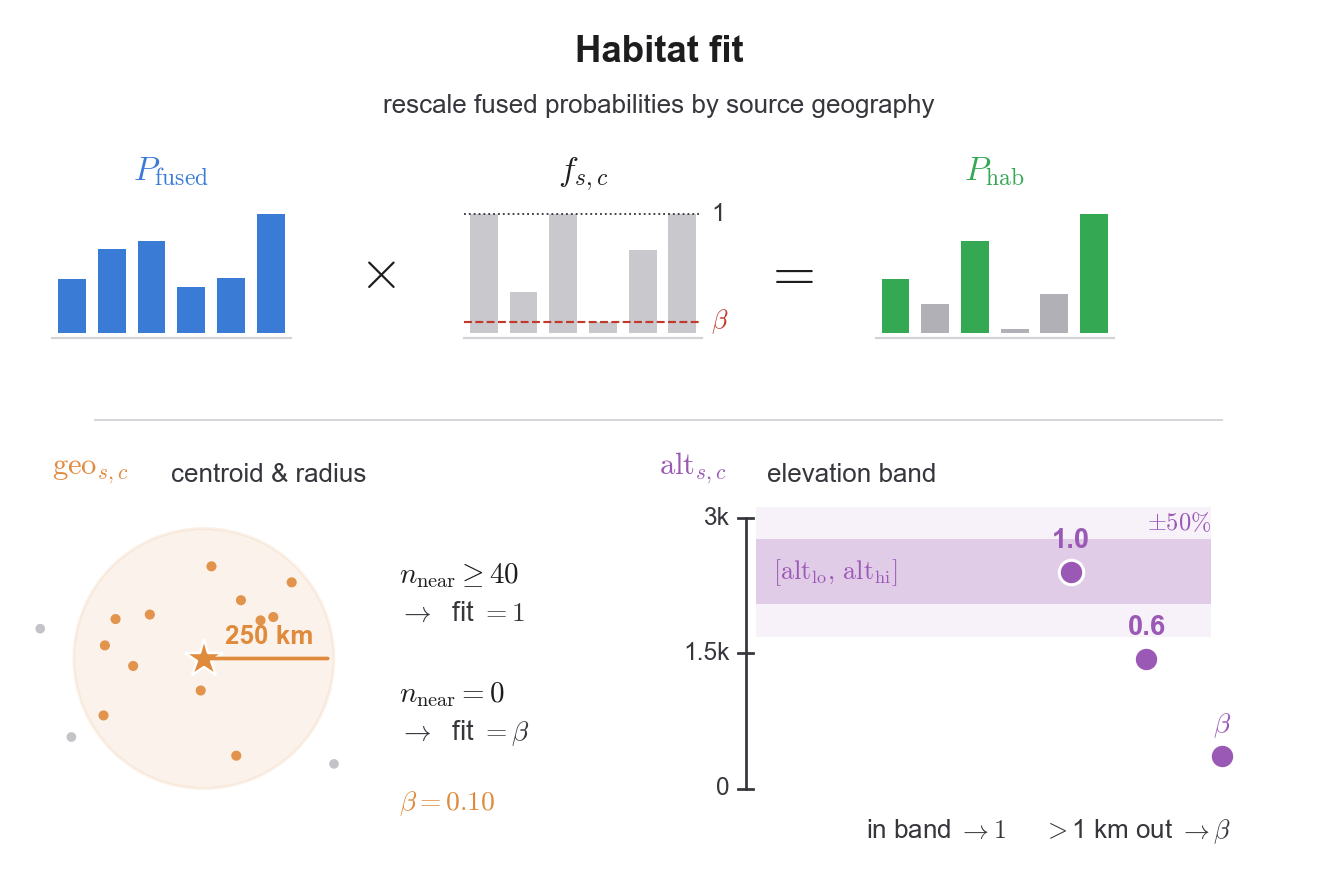}
    \caption{
    Habitat-fit demotion. The fused per-quadrat probabilities are rescaled by a source-dependent fit factor $f_{s,c}$ that encodes geographic and altitude priors. The geographic score reflects species presence within a 250\,km radius of the source centroid, while the altitude score penalizes species whose typical elevation lies outside the source-specific band.
    }
    \label{fig:habitat_fit_demotion}
\end{figure}

After classifier, kNN, and temporal fusion, the model produces a per-quadrat probability vector but does not explicitly account for the geographic distribution of species in the training data. The habitat-fit step injects this external prior.
For each quadrat $q$, the fused probabilities are rescaled using a source-dependent fit factor:
\begin{equation}
P_{\text{hab}}(q, c) = f_{S(q), c} \cdot P_{\text{fused}}(q, c),
\end{equation}
where $S(q)$ denotes the source of quadrat $q$. The fit factor is defined as the product of a geographic score and an altitude score:
\begin{equation}
f_{s,c} = \text{geo}_{s,c} \cdot \text{alt}_{s,c}.
\end{equation}

\paragraph{Geographic score.}
Each source $s$ is assigned a regional centroid $(\text{lat}_s, \text{lon}_s)$ computed as the median latitude and longitude of its anchor species (the top-30 most probable species per source, excluding cosmopolitan species that appear broadly across all sources); the centroid computation uses the publicly available PlantCLEF~2024 training metadata and is described in full in Appendix~\ref{app:habitat-fit}, which also tabulates the resulting values for all eight sources. For each species $c$, $n^{\text{near}}_{s,c}$ denotes the number of training observations within a radius $R = 250$ km of the centroid.
The geographic score is defined as:
\begin{equation}
\text{geo}_{s,c} = \beta + (1 - \beta) \cdot \min\left(1, \frac{n^{\text{near}}_{s,c}}{n^{\text{near}}_{\min}}\right),
\end{equation}
with $\beta = 0.10$ and $n^{\text{near}}_{\min} = 40$. Species with at least 40 nearby observations receive no penalty ($\text{geo}_{s,c} = 1$), while species with no nearby observations are downweighted to $\beta$.

\paragraph{Altitude score.}
Each source $s$ is associated with an altitude band $[\text{alt}^{\text{lo}}_s, \text{alt}^{\text{hi}}_s]$. For each species $c$, the median training-data altitude $\text{alt}_c$ is computed. If $\text{alt}_c$ lies within $\pm 50\%$ of the source band, we set $\text{alt}_{s,c} = 1$. Otherwise,
\begin{equation}
\text{alt}_{s,c} = \max\left(\beta, \; 1 - \frac{\text{dist}(\text{alt}_c, [\text{alt}^{\text{lo}}_s, \text{alt}^{\text{hi}}_s])}{1000}\right),
\end{equation}
where $\text{dist}(\cdot)$ denotes the distance (in meters) from $\text{alt}_c$ to the altitude band. Species whose altitude deviates by more than 1 km are clamped to $\beta$.

\subsection{Geographic Mask and Admission}

After habitat-fit demotion, a global South-Western Europe geographic mask is applied that removes approximately 3,000 of the 7,806 classes whose training-set latitude/longitude centroid falls outside the union of France, Spain, Italy, and Switzerland.
This mask follows the same strategy used in prior work~\cite{gustineli2025tile}.
The final per-quadrat probability vector is admitted using a two-step procedure:

\paragraph{1. Threshold + Top-$k$.}
All classes satisfying $P_{\text{hab}}(q, c) \geq 0.085$ are retained, and then truncate the result to the top-10 highest-probability classes.

\paragraph{2. Count-bound $N$-floor.}
For sources where the model tends to under-predict (e.g., \texttt{GUARDEN}, \texttt{LISAH}, \texttt{OPTMix}, \texttt{RNNB}), a minimum number of predictions is enforced. If fewer than $n_{\text{floor}} = 3$ species remain after Step 1, we admit additional candidates in descending probability order, subject to the constraint that each candidate satisfies a probability ratio
\begin{equation}
\frac{P_{\text{hab}}(q, c)}{\max_{c'} P_{\text{hab}}(q, c')} \geq r,
\end{equation}
with $r = 0.5$.
This mechanism allows confident near-ties to be included while preventing the admission of low-confidence predictions.

\subsection{TileQ-Decoder: Label-as-Query Multi-Label Decoding (Exploratory)}
\label{sec:tileq-decoder}
As an exploratory direction complementary to the aggregation-based main pipeline, a small ($\approx 4.5$M parameter) label-as-query transformer decoder, \emph{TileQ-Decoder}, was evaluated over per-tile CLS embeddings from the frozen organizer-provided ViTD2PC24All backbone on a single $4 \times 4$ tile grid.
The decoder follows the group-FFN design of ML-Decoder \cite{ridnik2023mldecoder}, with 64 query groups covering the 7,806 classes, and is trained with Asymmetric Loss \cite{ridnik2021asl} ($\gamma^- = 4$, $\gamma^+ = 0$, $m = 0.05$).

Because no real multi-label training data is available, synthetic ``pseudo-quadrats'' are constructed directly in CLS embedding space, exploiting the fact that the per-tile ViT forward pass has no cross-tile attention.
Each pseudo-quadrat is a $(16, 768)$ tensor of CLSs paired with a 7,806-dim multi-hot label.
Two samplers are compared under pre-registered hypotheses: \textbf{Sampler A} (naive) injects 5 species-CLSs from the training set and pads with 11 random-background train CLSs; \textbf{Sampler B} (test-anchored) injects 5 species-CLSs into 11 real test-quadrat tile CLSs, with FAISS top-1 retrieval ($\tau = 0.65$) used to relabel surviving backgrounds.
Two hypotheses were pre-registered before any Kaggle submission: \textbf{H1} that the decoder would beat the sorted-prob top-9 aggregator baseline ($F_1 = 0.30810$ on the same $4 \times 4$ tile grid), and \textbf{H2} that Sampler B would beat Sampler A by at least $+0.01$ macro-$F_1$.

\subsection{Cross-Region Transformer with Noisy-Student Distillation (Exploratory)}
\label{sec:cross-region}
As a second exploratory direction, a training-centric extension was evaluated based on a cross-region transformer trained with noisy-student distillation on the 212{,}782-image unlabeled LUCAS dataset \cite{dandrimont2022lucas}.
The approach extracts region-level embeddings from a frozen ViT and aggregates them with a lightweight transformer under pseudo-label supervision generated by a teacher model on the LUCAS imagery.
The outcome is discussed in Section \ref{sec:discussion}.

\subsection{SAM Instance-Crop Augmentation (Exploratory)}
\label{sec:sam-crops}
An inference-time augmentation using instance-aware crops from a SAM 3 segmentation model \cite{sam3} was also evaluated to supplement the grid tiles.
For each quadrat image, plants are segmented using SAM 3 with text prompts (``plant'' and ``flower'').
Detections are filtered by confidence score ($\geq 0.5$) and deduplicated with non-maximum suppression (IoU $\geq 0.6$).
Each remaining bounding-box crop is resized and encoded with the same ViT-L backbone, producing additional per-crop softmax distributions that are merged into the tile pool before aggregation.
The motivation is that grid tiles may straddle plant boundaries or contain too many plants, diluting strong indicators of certain plants.
Instance crops aligned to individual plants could improve recall for partially-visible or small specimens.

\section{Results}
\label{sec:results}

The final version of the pipeline achieves a macro-F1 score of $0.45777$ on the private leaderboard (public $0.46914$), adopted as the Baseline against which all ablations are measured. Three studies are reported: (A) a cumulative ablation that reduces the pipeline to a bare multi-scale classifier; (B) a multi-scale sweep that motivates the choice of scale set $S = \{3, 4, 5, 6\}$; and (C) a backbone comparison replacing ViT-L with ViT-B under matched multi-scale configurations.
This configuration corresponds to an intermediate run that was not designated as a final submission before the five-submission selection deadline; among the five selected submissions, the highest-scoring run reached private $F_1 = 0.43902$ (third place, public $0.51096$) (Table~\ref{tab:submitted_runs}).
Ablations are reported against the higher-scoring unselected configuration to isolate component contributions under the strongest version of the pipeline.

\subsection{Submitted Runs}
\label{sec:submitted-runs}

Table~\ref{tab:submitted_runs} lists all five official submissions alongside the paper baseline.
All five selected runs scored above $0.497$ on the public leaderboard, well above the paper baseline's public score of $0.46914$.
At submission time only the public score is visible; the paper baseline was therefore not identified as a top-performing run before the selection deadline, and its superior private performance was not observable until after the competition concluded.

\begin{table}[!htbp]
\centering
\caption{Official submissions and the paper baseline. Submission selection was based on the public F1 score alone (visible during the competition); private F1 is revealed post-competition. All five selected runs scored substantially higher on the public leaderboard than the paper baseline, explaining why the latter was not submitted despite achieving the highest private F1 of any pipeline run.}
\label{tab:submitted_runs}
\small
\begin{tabular}{l c c l}
\toprule
\textbf{Run} & \textbf{Private F1} & \textbf{Public F1} & \textbf{Status} \\
\midrule
Single pipeline (RNNB, $\beta{=}0.55$)        & 0.43902 & 0.51096 & Selected; 3\textsuperscript{rd} place \\
Single pipeline (OPTMix, $k{=}5$)             & 0.39844 & 0.49842 & Selected \\
Ensemble: per-source RRF + alpine blend        & 0.43553 & 0.52739 & Selected \\
Ensemble: interleaved merge                    & 0.43352 & 0.52387 & Selected \\
Ensemble: union merge                          & 0.43553 & 0.52739 & Selected \\
\midrule
Paper baseline ($3{+}4{+}5{+}6$, no ensemble) & \textbf{0.45777} & 0.46914 & Not submitted \\
\bottomrule
\end{tabular}
\end{table}

\subsection{Cumulative Ablation}

We remove pipeline components one at a time, in reverse order of their addition during development. Habitat-fit demotion, together with the ecological masks, produces the largest drop in performance ($-0.04075$ on the private leaderboard). The kNN retrieval blend is the second most impactful component ($-0.01256$), followed by the collapse to a single scale ($-0.01587$ when replacing the four-scale set with scale 5 alone). Temporal fusion (similarity-weighted aggregation on \texttt{RNNB} with per-source $\alpha$) yields the smallest contribution ($-0.00367$), but is retained due to its consistent gains across configurations.

\begin{table}[!htbp]
\centering
\caption{Cumulative ablation study. Components are removed sequentially from the full pipeline. Abl-5 is a non-cumulative backbone comparison.}
\label{tab:cumulative_ablation}
\begin{tabular}{l l c c c}
\toprule
\textbf{\#} & \textbf{Removes} & \textbf{Private F1} & \textbf{Public F1} & \textbf{$\Delta$ F1\_priv (from prev.\ row)} \\
\midrule
\textbf{Baseline} & -- & \textbf{0.45777} & \textbf{0.46914} & -- \\
\midrule
Abl-1 & kNN retrieval blend ($\beta{=}1.0$, no kNN) & 0.44521 & 0.44096 & -0.01256 \\
Abl-2 & habitat-fit demotion & 0.40446 & 0.40650 & -0.04075 \\
Abl-3 & temporal fusion & 0.40079 & 0.39835 & -0.00367 \\
Abl-4 & multi-scale aggregation ($3{+}4{+}5{+}6 \rightarrow 5$) & 0.38492 & 0.36886 & -0.01587 \\
\midrule
Abl-5$^{*}$ & ViT-L $\rightarrow$ ViT-B (matched $3{+}4{+}5{+}6$) & 0.36472 & 0.38672 & -0.03607 \\
\bottomrule
\end{tabular}
\end{table}

\subsection{Multiscale Tiling Sweep}

A range of multi-scale tiling configurations is evaluated under the Abl-3 setting (classifier-only, without kNN, temporal fusion, geographic masking, habitat-fit, or admission). Table~\ref{tab:multiscale_tiling} summarizes the key configurations around the final selection. The results show that combining multiple tiling scales consistently improves performance compared to using a single scale.

A clear pattern emerges across the table: larger scale sets (e.g., $4{+}5{+}6{+}7$, $4{+}5{+}6{+}7{+}8$) tend to perform better on the public leaderboard, whereas the private leaderboard peaks at $3{+}4{+}5{+}6$ and degrades as scales are added or removed. This divergence suggests mild overfitting to the public subset when using larger scale combinations.

\begin{table}[!htbp]
\centering
\caption{Multi-scale tiling configurations and corresponding performance for ViT-L classifier. The selected baseline configuration ($3{+}4{+}5{+}6$) is highlighted.}
\label{tab:multiscale_tiling}
\begin{tabular}{lccc}
\toprule
\textbf{Scales} & \textbf{Tiles / Image} & \textbf{Private F1} & \textbf{Public F1} \\
\midrule
3       & 9   & 0.38533 & 0.33787 \\
4       & 16  & 0.37920 & 0.35106 \\
5       & 25  & 0.38492 & 0.36886 \\
6       & 36  & 0.39332 & 0.39770 \\
7       & 49  & 0.38826 & 0.40523 \\
8       & 64  & 0.37075 & 0.41375 \\
\midrule
3+4+5   & 50  & 0.39865 & 0.38202 \\
4+5+6   & 77  & 0.39317 & 0.39882 \\
\textbf{3+4+5+6} & \textbf{86} & \textbf{0.40079} & \textbf{0.39835} \\
\midrule
3+4+5+6+7 & 135 & 0.39322 & 0.41064 \\
4+5+6+7   & 126 & 0.38961 & 0.41178 \\
4+5+6+7+8 & 190 & 0.37344 & 0.40219 \\
\bottomrule
\end{tabular}
\end{table}

\subsection{Backbone Comparison: ViT-B vs.\ ViT-L}

To isolate the contribution of the larger backbone, seven multi-scale configurations are re-evaluated using the ViT-B (\texttt{vit\_base\_patch14\_reg4\_dinov2.lvd142m}) checkpoint under the same Abl-4 setting.
For the ViT-B comparison, per-tile embeddings are extracted from the ViTD2PC24All checkpoint and fed into the same downstream pipeline (multi-scale tiling and max aggregation), with the only adjustment being the embedding dimension (768 for ViT-B vs. 1024 for ViT-L).

Two findings emerge. First, ViT-L consistently outperforms ViT-B across all tested configurations, with a private leaderboard gain ranging from $+0.022$ to $+0.052$ (mean $+0.039$). Second, the optimal multi-scale configuration is robust across backbones: both ViT-B and ViT-L achieve their best performance at $3{+}4{+}5{+}6$ on the private leaderboard. This indicates that the selected scale set is not specific to ViT-L, but rather reflects a property of the multi-scale aggregation under the test distribution.

\begin{table}[!htbp]
\centering
\caption{Backbone comparison between ViT-B and ViT-L across multi-scale configurations. ViT-L consistently outperforms ViT-B, with gains in private F1 across all settings.}
\label{tab:backbone_comparison}
\begin{tabular}{l c c c c}
\toprule
\textbf{Scales} & \textbf{Tiles / Image} & \textbf{ViT-B (priv / pub)} & \textbf{ViT-L (priv / pub)} & \textbf{$\Delta$ F1\_priv (L - B)} \\
\midrule
3       & 9   & 0.33377 / 0.35869 & 0.38533 / 0.33787 & +0.05156 \\
4       & 16  & 0.35749 / 0.34240 & 0.37920 / 0.35106 & +0.02171 \\
5       & 25  & 0.34103 / 0.36894 & 0.38492 / 0.36886 & +0.04389 \\
6       & 36  & 0.34651 / 0.36612 & 0.39332 / 0.39770 & +0.04681 \\
\midrule
4+5+6   & 77  & 0.35577 / 0.38887 & 0.39317 / 0.39882 & +0.03740 \\
\textbf{3+4+5+6} & \textbf{86} & \textbf{0.36472 / 0.38672} & \textbf{0.40079 / 0.39835} & \textbf{+0.03607} \\
4+5+6+7 & 126 & 0.34954 / 0.38866 & 0.38961 / 0.41178 & +0.04007 \\
\bottomrule
\end{tabular}
\end{table}

\subsection{TileQ-Decoder}
\label{sec:tileq-results}
Both pre-registered hypotheses were decisively rejected. Sampler A
reaches macro-$F_1 = 0.12180$ at top-$k = 9$ on the public leaderboard,
$0.186$ below the sorted-prob anchor and below even the team's no-grid
full-image baseline ($0.14506$); Sampler B reaches $0.09060$, $0.031$
below Sampler A in the opposite direction H2 anticipated.
A linear blend $s = \alpha \cdot \text{decoder} + (1 - \alpha) \cdot
\text{max\_prob}$ swept at $k = 10$ produces a monotonically decreasing
curve from $0.22144$ at $\alpha = 0$ to $0.11640$ at $\alpha = 1$,
confirming that the decoder probabilities are not even partially
redundant with the cached aggregator output.

\begin{table}[h]
    \caption{TileQ-Decoder results on the public leaderboard. The
    \textbf{anchor} is the sorted-prob top-9 aggregator on the same
    $4 \times 4$ per-tile probabilities (F1 = 0.30810). H1 rejection
    threshold was set at F1 $< 0.288$ before submission.}
    \label{tab:tileq}
    \centering
    \begin{tabular}{lcc}
        \toprule
        \textbf{Method} & \textbf{$k$} & \textbf{Public $F_1$} \\
        \midrule
        A0b: Sorted-prob top-9 (H1 anchor) & 9 & 0.30810 \\
        A1: TileQ-Decoder, Sampler A       & 9 & 0.12180 \\
        A2: TileQ-Decoder, Sampler B       & 9 & 0.09060 \\
        A3: $\alpha$-blend, $\alpha = 0$ (pure max\_prob) & 10 & 0.22144 \\
        A3: $\alpha$-blend, $\alpha = 1$ (pure decoder)   & 10 & 0.11640 \\
        \bottomrule
    \end{tabular}
\end{table}

\section{Discussion}
\label{sec:discussion}

The final system reaches $0.45777$ private F1 using a ViT-L backbone combined with several post-hoc components.
Habitat-fit is the dominant component in the pipeline. Removing the geographic score, altitude score, and ecological masks together results in a $-0.04075$ drop in private F1, significantly larger than any other ablation step. This behavior can be attributed to the nature of the information used. Habitat-fit incorporates geographic and altitude metadata from the training set, which is not directly accessible to the classifier at inference time. The prior is defined per source, with each source associated with its own geographic centroid and altitude band. This matches the structure of the test set and allows the adjustment to reflect source-specific species distributions.

The multi-scale sweep shows that performance is not monotonic with the number of scales.
In particular, adding scale 3 to $4{+}5{+}6{+}7$ does not improve results unless scale 7 is removed, with the best performance obtained at $3{+}4{+}5{+}6$.
This indicates that larger scale sets do not consistently add useful information and can instead introduce redundancy or noise at finer resolutions.
Conversely, single-scale and smaller combinations underperform, suggesting insufficient detail capture of the quadrat.
The selected configuration therefore reflects a balance between capturing fine-grained detail and maintaining stable tile-level predictions.
Consistent with this, larger scale sets tend to perform better on the public leaderboard but do not generalize to the private set, indicating that additional scales may overfit to the public subset rather than improve overall robustness.

ViT-L consistently outperforms ViT-B across all tested configurations, with gains between $+0.022$ and $+0.052$ private F1.
However, both backbones select the same optimal scale set ($3{+}4{+}5{+}6$).
This suggests that the choice of scales is determined primarily by the structure of the input images rather than backbone capacity.

Temporal fusion provides a relatively small gain ($-0.00367$) and is strongly source-dependent.
For RNNB, where vegetation varies significantly across visits, max-pooling tends to produce overconfident predictions dominated by a single visit; the similarity-weighted aggregation mitigates this by emphasizing visually consistent siblings.
Other sources such as alpine datasets exhibit low temporal variation and benefit from max-pooling, while single-visit sources do not support temporal aggregation at all.
As a result, applying the similarity-weighted approach outside RNNB consistently reduces performance.

A broader limitation of the pipeline is that several scalar hyperparameters (including the kNN blending weights ($\beta_{\text{multi}} = 0.70$, $\beta_{\text{single}} = 1.00$), the temporal fusion mixing coefficients ($\alpha_{\text{RNNB}} = 0.30$, $\alpha_{\text{default}} = 0.15$), the admission floor ($n_{\text{floor}} = 3$, $r = 0.5$), and the kNN temperature ($\tau = 0.07$)) were selected based on the public leaderboard score, which covers approximately 11\% of the test data. While the multi-scale results already demonstrate that performance on the public subset does not always generalise to the private set, the same risk applies to these post-hoc scalars: values that maximise the 11\% public signal may not be optimal over the full test distribution. We note that the habitat-fit priors and geographic mask are derived entirely from training-data statistics and are therefore not subject to this concern; the overfitting risk is specific to the admission and blending hyperparameters tuned against public leaderboard feedback.

The TileQ-Decoder null result (\S\ref{sec:tileq-results}) characterizes an under-explored failure mode for label-as-query decoding from CLS-domain pseudo-quadrats.
The dominant failure for Sampler B is what is termed \emph{FAISS-undersampling under-confidence}: because the empirical distribution of train$\leftrightarrow$test top-1 cosine similarities is unimodal at $\approx 0.51$ with no shoulder, only 5.46\% of test tiles receive a FAISS pseudo-label at $\tau = 0.65$.
The decoder therefore sees ``test-shaped tile $\rightarrow$ no species in label'' as the dominant supervision signal over 80k training steps, and at inference time outputs uniformly low logits, forcing top-$k$ decoding onto noise at the bottom of the distribution.
Sampler A's overconfident-but-train-shaped predictions retain enough mode-collapse onto plausible species to score higher on top-$k$ F1, despite its less principled supervision.
More broadly, the monotone-decreasing $\alpha$-blend indicates that for this task, aggregation-based methods that preserve the cached per-tile probability magnitudes substantially outperform synthetic-supervision approaches that re-process those features through a learned decoder.

The cross-region transformer extension (\S\ref{sec:cross-region}) did not improve performance.
The LUCAS dataset differs from the PlantCLEF  test distribution in both image characteristics and species composition,  so the teacher-generated pseudo-labels reflect this mismatch and the  student learns a distribution misaligned with the evaluation set.
Combined with the TileQ-Decoder result above, this reinforces a broader  pattern: for this task, training-centric extensions that introduce  distributional drift between training supervision and the test domain  tend to underperform inference-time strategies that operate directly on  the cached features of the in-domain fine-tuned classifier.

The SAM 3 instance-crop augmentation (\S\ref{sec:sam-crops}) yielded no consistent improvement relative to the corresponding runs with standard grid tiles.
This null result is hypothesized to be due to a distributional mismatch between the tiles and segmented crops.
The crops from the irregular segmentations occupy a different input space than the grid tiles on which both the ViT head and kNN index were calibrated, causing the added tiles to contribute noise rather than a discriminative signal.

\section{Conclusion}
\label{sec:conclusion}

This paper presents the DS@GT ARC third-place solution to the PlantCLEF 2026 multi-species plant identification challenge. The approach centers on a structured, inference-driven pipeline built around a fine-tuned DINOv2 ViT-L backbone, combining multi-scale tiling, retrieval-based ensembling, and source-aware post-processing.

The results show that the most impactful improvements arise from components that explicitly encode dataset structure rather than model complexity.
In particular, multi-scale aggregation enables robust detection of small and densely distributed species, while habitat-fit demotion and geographic masking provide strong gains by incorporating ecological priors not directly available to the classifier. 
Image similarity retrieval-based kNN ensembling and temporal fusion further improve performance, with the latter yielding consistent gains in multi-visit settings despite its relatively small overall contribution.

Training-centric extensions were also evaluated, including cross-region transformers with noisy-student distillation on the LUCAS dataset, but no performance gains were observed due to domain mismatch with the PlantCLEF test distribution.
An inference-time augmentation with instance-aware SAM3 crops similarly yielded no consistent improvement.
This highlights a key challenge of the task: effective solutions must align closely with the target distribution, and naive incorporation of external data or augmentations can degrade performance when this alignment is not preserved.

Overall, these findings suggest that for large-scale multi-species recognition under severe domain shift, carefully designed inference strategies and dataset-aware priors can outperform more complex training pipelines.
Future work may explore tighter integration between representation learning and ecological priors, as well as improved domain adaptation methods that better bridge the gap between single-label training data and multi-label field observations.
Separately, the exploration of a label-as-query transformer decoder over per-tile CLS embeddings (TileQ-Decoder) yielded a null result and a newly characterized failure mode for test-anchored synthetic supervision under sparse FAISS pseudo-labels, reinforcing the broader finding that for this task aggregation over cached per-tile probabilities is harder to beat than the methodological diversity of recent multi-label literature would suggest.

\begin{acknowledgments}
We thank the Data Science at Georgia Tech (DS@GT) ARC group for their support.
This research was supported in part through research cyberinfrastructure resources and services provided by the Partnership for an Advanced Computing Environment (PACE) at the Georgia Institute of Technology, Atlanta, Georgia, USA \cite{PACE}.
\end{acknowledgments}

\section*{Declaration on Generative AI}
During the preparation of this work, the author(s) used Claude in order to perform grammar and spelling check.
After using these tool(s)/service(s), the author(s) reviewed and edited the content as needed and take(s) full responsibility for the publication's content.


\bibliography{main}

\appendix

\section{Habitat-Fit Source Prior Computation}
\label{app:habitat-fit}

The habitat-fit factors used in Section~\ref{sec:methodology} are not hardcoded constants.
They are the output of a two-step computation performed once on the publicly available PlantCLEF~2024 training metadata (\texttt{PlantCLEF2024singleplanttrainingdata.csv}, provided by the challenge organizers) before inference begins.
The fit factors are then applied as a static lookup table at inference time, one vector per source.
All scripts are available in the public repository at \url{https://github.com/dsgt-arc/plantclef-2026}.

\paragraph{Step 1: Per-species geographic profile.}
For each of the 7,806 species in the training set, the script \texttt{build\_species\_geo\_profile.py} reads every training observation's latitude, longitude, and altitude from the metadata CSV.
For each source $s$ and each species $c$, it counts the number of training observations whose Haversine distance to source $s$'s centroid falls within the proximity radius $R = 250$~km.
This produces a per-species, per-source proximity count $n^{\text{near}}_{s,c}$, along with each species' median training altitude $\text{alt}_c$.
Species with no valid coordinate data receive a neutral fit factor of $1.0$.

\paragraph{Step 2: Refined source centroids.}
Initial hand-set centroids were replaced with data-derived centroids using \texttt{refine\_source\_centroids.py}.
For each of the eight test sources, the script identifies anchor species as the top-30 most frequently predicted species from an earlier locked pipeline stage, then removes cosmopolitan species (those appearing in the top-30 of three or more other sources).
The refined centroid is then computed as the median latitude and longitude of those anchor species' training observations, restricted to South-Western Europe (lat $35$°–$52$°N, lon $10$°W–$10$°E) to prevent tropical training images from pulling the centroid off-region.
These computed centroids replace the initial values and are stored in \texttt{refined\_source\_ref.json}.
Table~\ref{tab:source-priors} reports the resulting centroid coordinates and configured altitude bands for all eight sources.

\begin{table}[!htbp]
\centering
\caption{Per-source habitat-fit parameters used in the final submission pipeline.
Centroid coordinates (latitude, longitude) are computed from anchor-species training observations as described above; they are not hardcoded inputs.
Altitude bands encode the known ecological regime of each source and are used for the altitude-score term of the fit factor (Equation~10 in the main text).
Temporal fusion settings ($\alpha$, aggregator) are as described in Section~4.4.}
\label{tab:source-priors}
\small
\begin{tabular}{l c c c c c}
\toprule
\textbf{Source} & \textbf{Centroid (lat, lon)} & \textbf{Alt.\ band (m)} & \textbf{Ecology} & \textbf{Aggregator} & \textbf{$\alpha$} \\
\midrule
RNNB       & 43.21°N,\ 2.36°E  & 0–200   & Coastal sand dune        & Similarity-weighted & 0.30 \\
CBN-PdlC   & 45.21°N,\ 6.52°E  & 1500–3300 & Alpine (Pyrenees/Alps)  & Max                 & 0.15 \\
CBN-Pla    & 45.16°N,\ 6.60°E  & 1500–3300 & Alpine (Alps)           & Max                 & 0.15 \\
CBN-can    & 45.73°N,\ 5.97°E  & 900–2500  & Sub-alpine (Cantal)     & Max                 & 0.15 \\
GUARDEN    & 43.35°N,\ 2.89°E  & 0–200   & Mediterranean coastal    & Max                 & 0.15 \\
LISAH      & 45.00°N,\ 2.67°E  & 0–400   & Montpellier lowland      & Max                 & 0.15 \\
OPTMix     & 45.78°N,\ 2.81°E  & 50–800  & Central France mixed     & Max                 & 0.15 \\
2024-CEV3  & 46.23°N,\ 5.40°E  & 1500–3000 & Alpine                 & Max                 & 0.15 \\
\bottomrule
\end{tabular}
\end{table}

\noindent
The geographic score for species $c$ at source $s$ is then:
\[
\text{geo}_{s,c} = \beta + (1 - \beta) \cdot \min\!\left(1,\, \frac{n^{\text{near}}_{s,c}}{n^{\text{near}}_{\min}}\right),
\quad \beta = 0.10,\; n^{\text{near}}_{\min} = 40,
\]
and the altitude score follows Equation~12 in the main text.
The habitat-fit factor is $f_{s,c} = \text{geo}_{s,c} \cdot \text{alt}_{s,c}$.
Single-visit quadrats (approximately 24\% of the test set) receive no temporal fusion ($\alpha$ does not apply).

\section{Training Configuration}
\label{app:training}

The ViT-L/14 DINOv2 backbone is fine-tuned on the PlantCLEF~2024 single-plant training set (1.4M images, 7{,}806 species) using a three-phase recipe on the Partnership for an Advanced Computing Environment (PACE) cluster at Georgia Tech~\cite{PACE}.
All phases use 518$\times$518 input, ImageNet normalisation, AMP (float16), and random seed 42.
The ArcFace/LoRA retrieval head (Phase~3) is trained on the same dataset with the backbone frozen.
Table~\ref{tab:training-config} summarises the key hyperparameters.

\paragraph{Phase~1 (head warm-up).}
The DINOv2 backbone is frozen; only the 7{,}806-class linear classifier head is trained.
Optimiser: Adam, $\text{lr}=0.01$, no weight decay, cosine schedule with no warmup.
Batch size: 96 per GPU; up to 100 epochs with patience-based early stopping (patience~$=10$).
Augmentation: RandAugment (M$=$9), Mixup $\alpha{=}0.8$, CutMix $\alpha{=}1.0$, label smoothing $\epsilon{=}0.1$.

\paragraph{Phase~2 (full fine-tuning).}
Initialised from the Phase~1 best EMA checkpoint; all backbone parameters unfrozen.
Hardware: 4$\times$ H200 GPUs; effective batch $= 4\,\text{GPU}\times64\times4\,\text{grad.\ accum.}=1024$.
Optimiser: AdamW, peak $\text{lr}=8\times10^{-5}$, weight decay $=0.05$, layer-wise LR decay $=0.85$.
Scheduler: cosine, 2-epoch linear warmup, min lr $=10^{-7}$.
30 training epochs; drop-path $=0.3$; same Mixup/CutMix augmentation.
A 10-epoch \emph{cooldown} extension subsequently trains on the combined train$+$validation set at $\text{lr}=2\times10^{-5}$ with no warmup; the resulting EMA weights serve as the backbone for tile embedding extraction.

\paragraph{Phase~3 (ArcFace/LoRA retrieval head).}
The cooldown backbone is frozen.
LoRA adapters (rank~$r=16$, $\alpha=16$, applied to query/key/value projections of all 24 attention blocks) and a retrieval head ($1024\to1024\to\text{BN}\to\text{ReLU}\to1024\to\ell_2$-norm) are trained jointly with a sub-center ArcFace classifier ($K=3$ sub-centres, scale $s=30$, margin $m=0.30$).
Optimiser: AdamW, $\text{lr}=3\times10^{-4}$, weight decay $=10^{-4}$, 500-step cosine warmup, gradient clip $\|\cdot\|_2\leq5$.
Augmentation: Mixup $\alpha{=}0.8$, CutMix $\alpha{=}1.0$; batch size 64; 8 epochs.

\begin{table}[!htbp]
\centering
\caption{Training hyperparameters for the three fine-tuning phases.
All phases: AMP float16, EMA decay 0.9998, RandAugment M$=$9, Mixup $\alpha{=}0.8$, CutMix $\alpha{=}1.0$, label smoothing $\epsilon{=}0.1$, seed 42.}
\label{tab:training-config}
\small
\begin{tabular}{l c c c}
\toprule
\textbf{Setting} & \textbf{Phase~1} & \textbf{Phase~2} & \textbf{Phase~3} \\
 & \textit{head warm-up} & \textit{full fine-tune} & \textit{ArcFace/LoRA} \\
\midrule
Trainable params   & Head only        & All (${\approx}303$M)                & LoRA + head \\
Epochs             & $\leq\!100$ (pat.~10) & 30 $+$ 10 cooldown          & 8 \\
Optimizer          & Adam             & AdamW                                & AdamW \\
Peak LR            & $1\times10^{-2}$ & $8\times10^{-5}$                     & $3\times10^{-4}$ \\
Weight decay       & 0                & 0.05                                 & $10^{-4}$ \\
Layer-wise LR decay & —               & 0.85                                 & — \\
Effective batch    & 96               & 1024 (4 GPU $\times$ 64 $\times$ 4 acc.) & 64 \\
LR schedule        & Cosine           & Cosine, 2-ep warmup                  & Cosine, 500-step warmup \\
Drop path          & 0.05             & 0.30                                 & — \\
LoRA rank          & —                & —                                    & 16 \\
ArcFace $(K,s,m)$  & —                & —                                    & $(3,\,30,\,0.30)$ \\
Hardware           & PACE H200        & 4$\times$ PACE H200                  & PACE H200 \\
\bottomrule
\end{tabular}
\end{table}

\end{document}